\documentclass[3p]{elsarticle}

\usepackage[utf8]{inputenc}
\usepackage[T1]{fontenc}
\usepackage{hyperref}
\usepackage{url}
\usepackage{booktabs}
\usepackage{multirow}
\usepackage{graphicx}
\usepackage{float}
\usepackage{amsfonts}
\usepackage{amsmath}
\usepackage{algorithm}
\usepackage{algorithmic}
\usepackage{caption}
\usepackage{subcaption}
\usepackage{xcolor}
\usepackage{xurl}
\usepackage{comment}

\DeclareMathOperator*{\argmax}{argmax}

\makeatletter
\def\ps@pprintTitle{%
 \let\@oddhead\@empty
 \let\@evenhead\@empty
 \def\@oddfoot{}%
 \let\@evenfoot\@oddfoot}
\makeatother

\begin{document}

\begin{frontmatter}

\title{Risk-Sensitive Policy with Distributional Reinforcement Learning}
\author[1]{Thibaut Théate}\corref{cor1}
\ead{thibaut.theate@uliege.be}
\author[1,2]{Damien Ernst}
\ead{dernst@uliege.be}
\address[1]{Department of Electrical Engineering and Computer Science, University of Liège, Liège, Belgium}
\address[2]{Information Processing and Communications Laboratory, Institut Polytechnique de Paris, Paris, France}
\cortext[cor1]{Corresponding author.}

\begin{abstract}

    Classical reinforcement learning (RL) techniques are generally concerned with the design of decision-making policies driven by the maximisation of the expected outcome. Nevertheless, this approach does not take into consideration the potential risk associated with the actions taken, which may be critical in certain applications. To address that issue, the present research work introduces a novel methodology based on distributional RL to derive sequential decision-making policies that are sensitive to the risk, the latter being modelled by the tail of the return probability distribution. The core idea is to replace the $Q$ function generally standing at the core of learning schemes in RL by another function taking into account both the expected return and the risk. Named the \textit{risk-based utility function} $U$, it can be extracted from the random return distribution $Z$ naturally learnt by any distributional RL algorithm. This enables to span the complete potential trade-off between risk minimisation and expected return maximisation, in contrast to fully risk-averse methodologies. Fundamentally, this research yields a truly practical and accessible solution for learning risk-sensitive policies with minimal modification to the distributional RL algorithm, and with an emphasis on the interpretability of the resulting decision-making process.

\end{abstract}


\end{frontmatter}

\section{Introduction}
\label{SectionIntroduction}

The reinforcement learning (RL) approach is concerned with the learning process of a sequential decision-making policy based on the interactions between an agent and its environment \cite{Sutton2018}. More precisely, the training is based on the rewards acquired by the agent from the environment as a consequence of its actions. Within this context, the objective is to identify the actions maximising the discounted sum of rewards, also named return. There exist multiple sound approaches based on the RL paradigm, and key successes/milestones have been achieved throughout the years of research. Nevertheless, these RL algorithms mostly rely on the expectation of the return, not on its complete probability distribution. For instance, the popular \textit{Q-learning} methodology is based on the modelling of the $Q$ function, which can in fact be seen as an estimation of the expected return \cite{Watkins1992}.\\

While focusing exclusively on the expectation of the return has already proven to be perfectly sound for numerous applications, this approach however exhibits clear limitations for other decision-making problems. Indeed, some areas of application may also require to properly mitigate the risk associated with the actions taken \cite{Paduraru2021}. One could for instance mention the healthcare \cite{Gottesman2019} and finance \cite{Theate2020} sectors, but also robotics in general \cite{Thananjeyan2021}, and especially autonomous driving \cite{Zhu2022}. Such a requirement for risk management may not only be true for the decision-making policy, but potentially also for the exploration policy during the learning process. In addition, properly taking into consideration the risk may be particularly convenient in environments characterised by substantial uncertainty.\\

The present research work suggests to take advantage of the distributional RL approach, belonging to the \textit{Q-learning} category, in order to learn risk-sensitive policies. Basically, a distributional RL algorithm targets the complete probability distribution of the random return rather than only its expectation \cite{Bellemare2017}. This methodology presents key advantages. Firstly, it enables the learning of a richer representation of the environment, which may lead to an increase in the decision-making policy performance. Secondly, the distributional RL approach contributes to improve the explainability of the decision-making process, which is key in machine learning to avoid black-box models. Lastly and most importantly for this research work, it makes possible the convenient derivation of decision-making policies but also exploration strategies that are sensitive to the risk.\\

The core idea promoted by this research work is the use of the \textit{risk-based utility function} $U$ as replacement of the popular $Q$ function for action selection. In fact, it may be seen as an extension of the $Q$ function taking into consideration the risk, which is assumed to be represented by the worst returns achievable by a policy. Therefore, the function $U$ is to be derived from the complete probability distribution of the random return $Z$, which is learnt by any distributional RL algorithm. The single modification to that RL algorithm to learn risk-sensitive policies is to employ the utility function $U$ rather than the expected return $Q$ for both exploration and decision-making. This allows the presented approach to become a very practical and interpretable RL solution for achieving risk-sensitive decision-making.

\section{Literature review}
\label{SectionLiterature}

The core objective of the classical RL approach is to learn optimal decision-making policies without any concerns about the risk or safety \cite{Sutton2018}. The resulting policies are said to be \textit{risk-neutral}. Nevertheless, there are numerous real-world applications requiring to take into consideration the risk in order to ensure safer decision-making \cite{Paduraru2021}. Two main approaches can be identified for achieving safe RL. Firstly, the optimality criterion can be modified so that a safety factor is included. Secondly, the exploration process can be altered based on a risk metric \cite{Garcia2015}. These techniques give rise to \textit{risk-sensitive} or \textit{risk-averse} policies.\\

Scientific research on risk-sensitive RL has been particularly active for the past decade. Various relevant risk criteria have been studied for that purpose. The most popular ones are undoubtedly the \textit{mean-variance} \cite{Castro2012, Prashanth2013, Zhang2021} and the \textit{(Conditional) Value at Risk} (CVaR) \cite{Rockafellar2001, Chow2015, Chow2017}. Innovative techniques have been introduced for both \textit{policy gradient} \cite{Tamar2015, Rajeswaran2017, Hiraoka2019} and \textit{value iteration} \cite{Shen2014, Dabney2018, Tang2019, Urpi2021, Yang2022} approaches, with the solutions proposed covering both discrete and continuous action spaces. Additionally, risk-sensitive methodologies have also been studied in some niche sub-fields of RL, such as robust adversarial RL \cite{Pinto2017}, but also multi-agent RL \cite{Qiu2021}.\\

Focusing on the \textit{value iteration} methodology, the novel distributional RL approach \cite{Bellemare2017} has been a key breakthrough, by giving access to the full probability distribution of the random return. For instance, \cite{Dabney2018} suggests to achieve risk-sensitive decision-making via a distortion risk measure. Applied on top of the IQN distributional RL algorithm, this is in fact equivalent to changing the sampling distribution of the quantiles. In \cite{Tang2019}, a novel actor-critic framework is presented, based on the distributional RL approach for the critic component. The latter work is extended to the offline setting in \cite{Urpi2021}, since training RL agents online may be prohibitive because of the risk inevitably induced by exploration. One can finally mention \cite{Yang2022} that introduces the \textit{Worst-Case Soft Actor Critic} (WCSAC) algorithm, which is based on the approximation of the probability distribution of accumulated safety-costs in order to achieve risk control. More precisely, a certain level of CVaR, estimated from the distribution, is regarded as a safety constraint.\\

In light of this literature, the novel solution introduced in this research paper presents key advantages. Firstly, the methodology proposed is relatively simple and can be applied on top of any distributional RL algorithm with minimal modification to the core algorithm. Secondly, the proposed approach enables to span the entire potential trade-off between risk minimisation and expected return maximisation. According to the user's needs, the policy learnt can be risk-averse, risk-neutral or in between the two (risk-sensitive). Lastly, the solution presented contributes to improve the interpretability of the decision-making policy learnt.

\section{Theoretical background}
\label{SectionBackground}

\subsection{Markov decision process}
\label{SectionMDP}

Traditionally in RL, the interactions between the agent and its environment are modelled as a \textit{Markov decision process} (MDP). An MDP is a 6-tuple $(\mathcal{S},\ \mathcal{A},\ p_R,\ p_T,\ p_0,\ \gamma)$ where $\mathcal{S}$ and $\mathcal{A}$ respectively are the state and action spaces, $p_R(r|s,a)$ is the probability distribution from which the reward $r \in \mathbb{R}$ is drawn given a state-action pair $(s, a)$, $p_T(s'|s,a)$ is the transition probability distribution, $p_0(s_0)$ is the probability distribution over the initial states $s_0 \in \mathcal{S}$, and $\gamma \in [0, 1[$ is the discount factor. The RL agent makes decisions according to its policy $\pi: \mathcal{S} \rightarrow \mathcal{A}$, assumed deterministic, mapping the states $s \in \mathcal{S}$ to the actions $a \in \mathcal{A}$.

\subsection{Distributional reinforcement learning}
\label{SectionDistributionalRL}

In classical \textit{Q-learning} RL, the core idea is to model the \textit{state-action value function} $Q^{\pi} : \mathcal{S} \times \mathcal{A} \rightarrow \mathbb{R}$ of a policy $\pi$. This important quantity $Q^{\pi}(s, a)$ represents the expected discounted sum of rewards obtained by executing an action $a \in \mathcal{A}$ in a state $s \in \mathcal{S}$ and then following a policy $\pi$:
\begin{equation}
    Q^{\pi}(s, a) = \underset{s_t, r_t}{\mathbb{E}} \left[\sum_{t=0}^{\infty} \gamma^{t} r_t\right] \text{,} \ \ \ \ \ (s_0, a_0) := (s, a),\ a_{t} = \pi(s_{t}) \text{.}
\end{equation}

Key to the learning process is the \textit{Bellman equation} \cite{Bellman1957}, that the $Q$ function satisfies:
\begin{equation}
    Q^{\pi}(s, a) = \underset{s', r}{\mathbb{E}} \left[r + \gamma Q^{\pi}(s', \pi(s'))\right] \text{.}
\end{equation}

In classical RL, the main objective is to determine an \textit{optimal policy} $\pi^*$ which can be defined based on the \textit{optimal state-action value function} $Q^* : \mathcal{S} \times \mathcal{A} \rightarrow \mathbb{R}$ as follows:
\begin{equation}
    Q^{*}(s, a) = \underset{s', r}{\mathbb{E}} \left[r + \gamma \max_{a' \in \mathcal{A}}Q^{*}(s', a')\right] \text{,}
\end{equation}
\begin{equation}
    \pi^*(s) \in \argmax_{a \in \mathcal{A}} \ Q^*(s, a)\ \text{.}
\end{equation}

The optimal policy $\pi^*$ maximises the expected return (discounted sum of rewards). This research work later presents an alternative objective criterion for optimality in a risk-sensitive RL setting.\\

\begin{figure}[H]
    \centering
    \includegraphics[width=0.7\linewidth, trim={7cm 7.7cm 7.7cm 7.3cm}, clip]{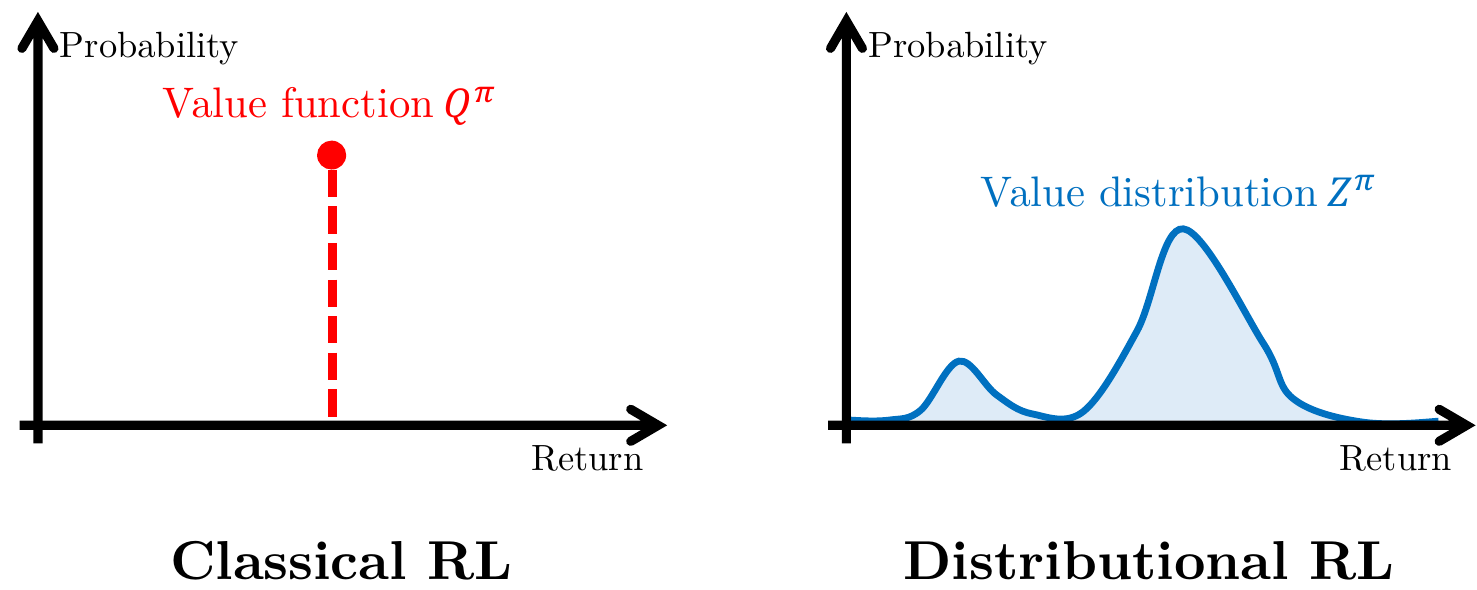}
    \caption{Intuitive graphical comparison between classical RL and distributional RL, for a decision-making policy $\pi$.}
    \label{FigureDistributionalRL}
\end{figure}

The distributional RL approach goes a step further by modelling the complete probability distribution over returns instead of only its expectation \cite{Bellemare2017}, as illustrated in Figure \ref{FigureDistributionalRL}. To this end, let the reward $R(s, a)$ be a random variable distributed under $p_R(\cdot|s, a)$, the \textit{state-action value distribution} $Z^{\pi} \in \mathcal{Z}$ of a policy $\pi$ is a random variable defined as follows:
\begin{equation}
    Z^{\pi}(s, a) \stackrel{D}{=} \sum_{t=0}^{\infty} \gamma^{t} R(s_t, a_t)\ \text{,} \ \ \ \ \ (s_0, a_0) := (s, a),\ a_{t} = \pi(s_{t}),\ s_{t+1} \sim p_T(\cdot|s_t, a_t)\ \text{,}
\end{equation}

where $A \stackrel{D}{=} B$ denotes the equality in probability distribution between the random variables $A$ and $B$. Therefore, the state-action value function $Q^{\pi}$ is the expectation of the \textit{random return} $Z^{\pi}$. Equivalently to the expected case, there exists a \textit{distributional Bellman equation} that recursively describes the random return $Z^{\pi}$ of interest:
\begin{equation}
\label{EquationDistributionalBellman}
    Z^{\pi}(s, a) \stackrel{D}{=} R(s, a) + \gamma P^{\pi} Z^{\pi}(s, a)\ \text{,}
\end{equation}
\begin{equation}
    P^{\pi} Z^{\pi}(s, a) :\stackrel{D}{=} Z^{\pi}(s', a')\ \text{,} \ \ \ \ \ s' \sim p_T(\cdot|s, a),\ a' = \pi(s')\  \text{,}
\end{equation}
where $P^{\pi} : \mathcal{Z} \rightarrow \mathcal{Z}$ is the transition operator. To end this section, one can define the \textit{distributional Bellman operator} $\mathcal{T}^{\pi} : \mathcal{Z} \rightarrow \mathcal{Z}$ together with the \textit{distributional Bellman optimality operator} $\mathcal{T}^* : \mathcal{Z} \rightarrow \mathcal{Z}$ as follows:
\begin{equation}
    \mathcal{T}^{\pi} Z^{\pi}(s, a) \stackrel{D}{=} R(s, a) + \gamma P^{\pi} Z^{\pi}(s, a)\ \text{,}
\end{equation}
\begin{equation}
    \mathcal{T}^* Z^*(s, a) \stackrel{D}{=} R(s, a) + \gamma Z^*\left(s', \pi^*(s')\right) \text{,} \ \ \ \ \ s' \sim p_T(\cdot|s, a)\ \text{.}
\end{equation}

\vspace{0.3cm}

A distributional RL algorithm may be characterised by two core features. Firstly, both the representation and parameterisation of the random return probability distribution have to be properly selected. There exists multiple solutions for representing a unidimensional distribution: probability density function (PDF), cumulative distribution function (CDF), quantile function (QF). In practice, deep neural networks (DNNs) are generally used for the approximation of these particular functions. The second fey feature relates to the probability metric adopted for comparing two distributions, such as the Kullback-Leibler (KL) divergence, the Cramer distance or the Wasserstein distance. More precisely, the role of the probability metric in distributional RL is to quantitatively compare two probability distributions of the random return so that a temporal difference (TD) learning method is applied, in a similar way to the mean squared error between Q-values in classical RL. The probability metric plays an even more important role as different metrics offer distinct theoretical convergence guarantees for distributional RL.

\section{Methodology}
\label{SectionMethodology}

\subsection{Objective criterion for risk-sensitive RL}
\label{SectionObjective}

As previously explained, the objective in classical RL is to learn a decision-making policy $\pi \in \Pi$ that maximises in expectation the discounted sum of rewards. Formally, this objective criterion can be expressed as the following:
\begin{equation}
    \label{EquationObjectiveRL}
    \underset{\pi}{\text{maximise}}\ \mathbb{E} \left[ \sum_{t=0}^{\infty} \gamma^t r_t \right] \text{.}
\end{equation}

\vspace{0.3cm}

In order to effectively take into consideration the risk and value its mitigation, this research work presents an update of the former objective. In fact, coming up with a generic definition for the risk is not trivial since the risk is generally dependent on the decision-making problem itself. In the present research work, it is assumed that the risk is assessed on the basis of the worst returns achievable by a policy $\pi$. Therefore, a successful decision-making policy should ideally maximise the expected discounted sum of rewards while avoiding low values for the worst case returns. The latter requirement is approximated with a new constraint attached to the former objective defined in Equation \ref{EquationObjectiveRL}: the probability of having the policy achieving a return lower than a certain minimum value should not exceed a given threshold. Mathematically, the alternative objective criterion proposed for risk-sensitive RL can be expressed as follows:
\begin{equation}
\label{EquationObjectiveRLRisk}
\begin{aligned}
& \underset{\pi}{\text{maximise}}
& & \mathbb{E} \left[ \sum_{t=0}^{\infty} \gamma^t r_t \right] \text{,}\\
& \text{such that}
& & p \left[ \sum_{t=0}^{\infty} \gamma^t r_t \le R_{\text{min}} \right] \le \epsilon \ \text{,}
\end{aligned}
\end{equation}
\newpage
where:
\begin{itemize}
    \item [$\bullet$] $p[\star]$ denotes the probability of the event $\star$,
    \item [$\bullet$] $R_{\text{min}}$ is the minimum acceptable return (from the perspective of risk mitigation),
    \item [$\bullet$] $\epsilon \in [0, 1]$ is the threshold probability to not exceed.
\end{itemize}

\subsection{Practical modelling of the risk}
\label{SectionRisk}

\begin{figure}[b]
    \centering
    \includegraphics[width=0.6\linewidth, trim={8.5cm 7.3cm 7.8cm 6.7cm}, clip]{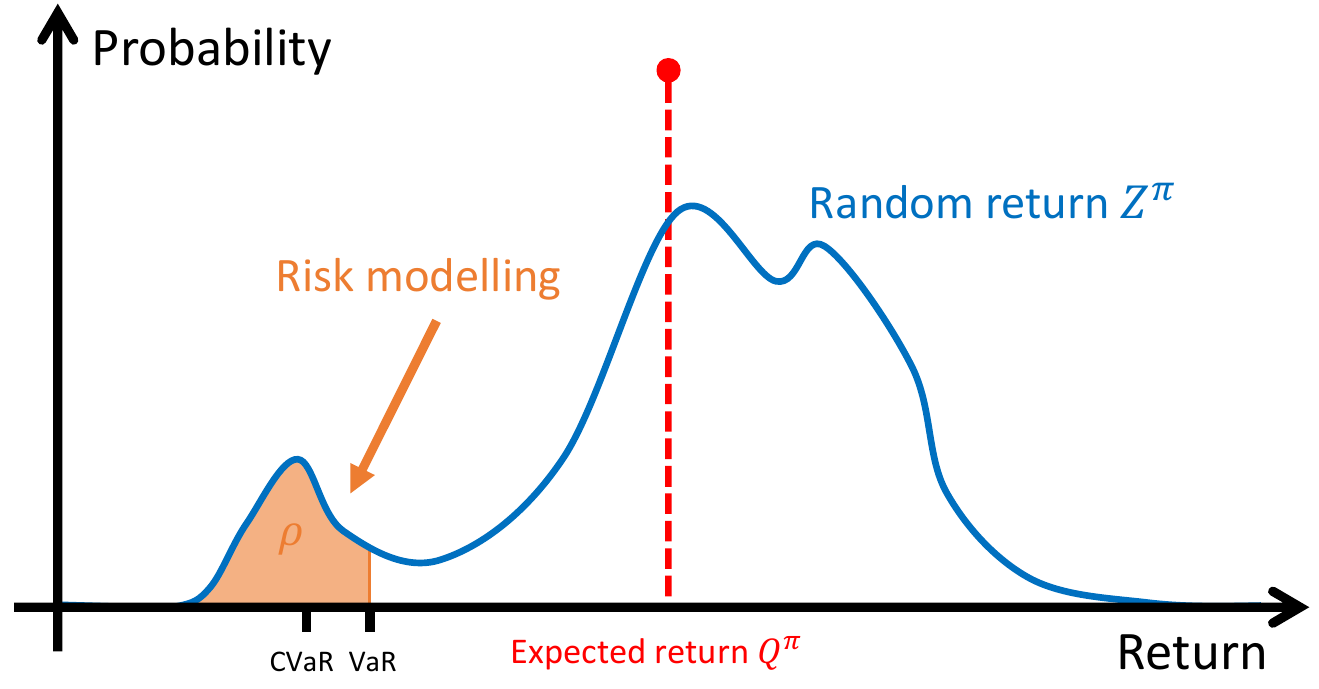}
    \caption{Illustration of the risk modelling adopted in this research work, based on the probability distribution of the random return $Z^{\pi}$ learnt by a distributional RL algorithm.}
    \label{FigureRiskModelling}
\end{figure}

As previously hinted, this research work assumes that the risk associated with an action is related to the worst achievable returns when executing that particular action and then following a certain decision-making policy $\pi$. In such a context, the distributional RL approach becomes particularly interesting, by providing access to the full probability distribution of the random return $Z^{\pi}$. Thus, the risk can be efficiently assessed by examining the so-called tail of the learnt probability distribution. Moreover, the new constraint in Equation \ref{EquationObjectiveRLRisk} can be approximated through popular risk measures such as the \textit{Value at Risk} and \textit{Conditional Value at Risk}. Illustrated in Figure \ref{FigureRiskModelling}, these two risk measures are formally expressed as follows:\\
\begin{equation}
    \text{VaR}_{\rho}(Z^{\pi}) = \text{inf} \left\{z \in \mathbb{R} : F_{Z^{\pi}}(z) \ge \rho \right\} \text{,}
\end{equation}
\begin{equation}
    \text{CVaR}_{\rho}(Z^{\pi}) = \mathbb{E} \left[z \ | \ z \le \text{VaR}_{\rho}(Z^{\pi}) \right] \text{,}
\end{equation}
where $F_{Z^{\pi}}$ represents the CDF of the random return $Z^{\pi}$.\\

More generally, this research work introduces the \textit{state-action risk function} $R^{\pi} : \mathcal{S} \times \mathcal{A} \rightarrow \mathbb{R}$ of a decision-making policy $\pi$, which is the equivalent of the $Q$ function for the risk. More precisely, that function $R^{\pi}(s, a)$ quantifies the riskiness of the discounted sum of rewards obtained by executing an action $a \in \mathcal{A}$ in a state $s \in \mathcal{S}$ and then following a policy $\pi$:
\begin{equation}
    R^{\pi}(s, a) = \mathcal{R}_{\rho} \left[ Z^{\pi}(s, a) \right] \text{,}
\end{equation}
where:
\begin{itemize}
    \item [$\bullet$] $\mathcal{R}_{\rho} : \mathcal{Z} \rightarrow \mathbb{R}$ is a function extracting risk features from the random return probability distribution $Z^{\pi}$, such as $\text{VaR}_{\rho}$ or $\text{CVaR}_{\rho}$,
    \item [$\bullet$] $\rho \in \left]0, 1\right[$ is a parameter corresponding to the cumulative probability associated with the worst returns, generally between $0\%$ and $10\%$. In other words, this parameter controls the size of the random return distribution tail from which the risk is estimated.
\end{itemize}

\subsection{Risk-based utility function}
\label{SectionUtilityFunction}

\begin{figure}[b]
    \centering
    \includegraphics[width=0.6\linewidth, trim={8.5cm 7.3cm 7.8cm 6.7cm}, clip]{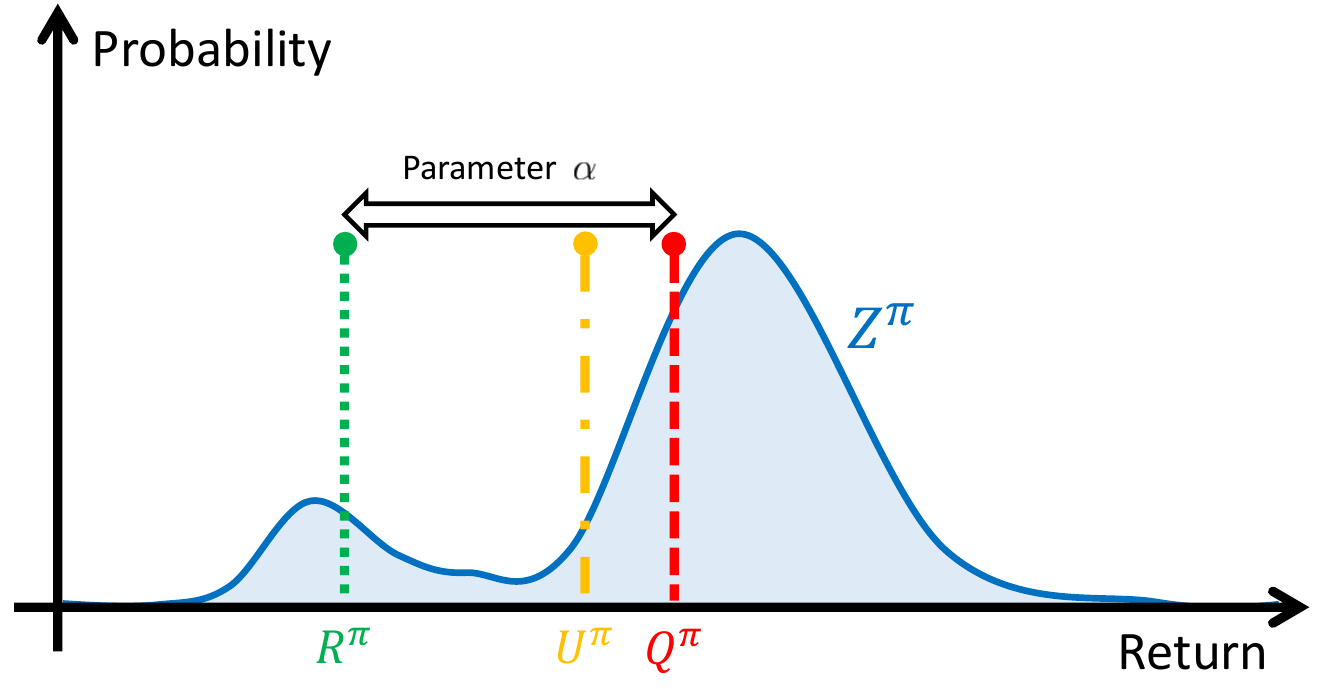}
    \caption{Illustration of the utility function $U^{\pi}$ for a typical random return probability distribution, with $\alpha = 0.75$ in this case.}
    \label{FigureUtilityFunction}
\end{figure}

In order to pursue the objective criterion defined in Equation \ref{EquationObjectiveRLRisk} for risk-sensitive RL, this research work introduces a new concept: the \textit{state-action risk-based utility function}. Denoted $U^{\pi}(s, a)$, the utility function assesses the quality of an action $a \in \mathcal{A}$ in a certain state $s \in \mathcal{S}$, in terms of expected performance and risk, assuming that the policy $\pi$ is followed afterwards. In fact, the intent is to extend the popular $Q$ function so that the risk is taken into consideration, by taking advantage of the risk function defined in Section \ref{SectionRisk}. More precisely, the utility function $U^{\pi}$ is built as a linear combination of the $Q^{\pi}$ and $R^{\pi}$ functions. Formally, the risk-based utility function $U^{\pi} : \mathcal{S} \times \mathcal{A} \rightarrow \mathbb{R}$ of a policy $\pi$ is defined as the following:
\begin{align}
    \label{EquationUtilityFunction}
    U^{\pi}(s, a)
    &= \alpha \ Q^{\pi}(s, a) \ + \ (1 - \alpha) \ R^{\pi}(s, a) \\
    &= \alpha \ \mathbb{E}\left[ Z^{\pi}(s, a) \right] \ + \ (1 - \alpha) \ \mathcal{R}_\rho \left[ Z^{\pi}(s, a) \right] \text{,}
\end{align}
where $\alpha \in \left[0, 1\right]$ is a parameter controlling the trade-off between expected performance and risk. If $\alpha = 0$, the utility function will be maximised with a fully risk-averse decision-making policy. On the contrary, if $\alpha = 1$, the utility function degenerates into the $Q$ function quantifying the performance on expectation. Figure \ref{FigureUtilityFunction} graphically describes the utility function $U^{\pi}$, which moves along the x-axis between the quantities $R^{\pi}$ and $Q^{\pi}$ when modifying the value of the parameter $\alpha$.

\subsection{Risk-sensitive distributional RL algorithm}
\label{SectionAlgorithm}

\begin{algorithm*}
\caption{Risk-sensitive distributional RL algorithm}
\begin{algorithmic}
\STATE Initialise the experience replay memory $M$ of capacity $C$.
\STATE Initialise both the main and target DNN weights $\theta = \theta^-$.
\FOR{episode = 0 \TO $N$}
    \FOR{time step $t = 0$ \TO $T$, or until episode termination}
    \STATE Acquire the state $s$ from the environment $\mathcal{E}$.
        \STATE With probability $\epsilon$, select a random action $a \in \mathcal{A}$.
        \STATE Otherwise, select the action $a = \argmax_{a' \in \mathcal{A}} \textcolor{red}{U^{\pi}(s, a'; \theta)}$.
        \STATE Execute action $a$ in environment $\mathcal{E}$ to get the next state $s'$ and the reward $r$.
        \STATE Store the experience $e = (s, a, r, s')$ in $M$.
        \STATE Randomly sample from $M$ a minibatch of $N_e$ experiences $e_i = (s_i, a_i, r_i, s_{i}^{'})$.
        \FOR{$i = 0$ \TO $N_e$}
            \STATE Distributional Bellman equation: $Z^{\pi}(s_i, a_i; \theta) \stackrel{D}{=} r_i + \gamma Z^{\pi}(s_{i+1}, \argmax_{a_i' \in \mathcal{A}} \textcolor{red}{U^{\pi}(s_{i+1}, a_i'; \theta^-)}; \theta^-)$.
        \ENDFOR
        \STATE Compute the resulting loss $\mathcal{L}(\theta)$, according to the probability metric selected.
        \STATE Update the main DNN parameters $\theta$ using a deep learning optimiser with learning rate $L_r$.
        \STATE Update the target DNN parameters $\theta^- = \theta$ every $N^-$ steps.
        \STATE Anneal the $\epsilon$-greedy exploration parameter $\epsilon$.
    \ENDFOR
\ENDFOR
\end{algorithmic}
\label{AlgorithmRiskSensitive}
\end{algorithm*}

In most applications, the motivation for choosing the distributional RL approach over the classical one is related to the improved expected performance that results from the learning of a richer representation of the environment. Despite having access to the full probability distribution of the random return, only the expectation is exploited to derive decision-making policies:
\begin{equation}
\label{EquationExpectationPolicy}
    \pi(s) \in \argmax_{a \in \mathcal{A}}\ \underbrace{\mathbb{E} \left[ Z^{\pi}(s, a) \right]}_{Q^{\pi}(s, a)} \text{.}
\end{equation}

Nevertheless, as previously hinted, the random return $Z^{\pi}$ does also contain valuable information about the risk, which could be exploited to learn risk-sensitive decision-making and exploration policies. The present research work suggests to achieve risk-sensitive distributional RL by maximising the utility function $U^{\pi}$, derived from $Z^{\pi}$, instead of the expected return $Q^{\pi}$ when selecting actions. This alternative operation would be performed during both exploration and exploitation. Even though maximising the utility function is not exactly equivalent to the optimisation of the objective criterion defined in Equation \ref{EquationObjectiveRLRisk}, it is a relevant step towards risk-sensitive RL. Consequently, a risk-sensitive policy $\pi$ can be derived as follows:
\begin{equation}
\label{EquationRiskSensitivePolicy}
    \pi(s) \in \argmax_{a \in \mathcal{A}}\ U^{\pi}(s, a) \ \text{.}
\end{equation}

Throughout the learning phase, a classical Q-learning algorithm is expected to progressively converge towards the optimal value function $Q^*$ that naturally arises from the optimal policy $\pi^*$. In a similar way, the proposed risk-sensitive RL algorithm jointly learns the optimal policy $\pi^*$ and the \textit{optimal state-action risk-based utility function} $U^*$. More formally, the latter two are mathematically defined as the following:
\begin{equation}
    U^{*}(s, a) = \alpha \ \mathbb{E}\left[ Z^{*}(s, a) \right] \ + \ (1 - \alpha) \ \mathcal{R}_\rho \left[ Z^{*}(s, a) \right] \text{,}
\end{equation}
\begin{equation}
    Z^{*}(s, a) \stackrel{D}{=} R(s, a) + \gamma Z^*\left(s', \pi^*(s')\right) \text{,}
\end{equation}
\begin{equation}
    \pi^*(s) \in \argmax_{a \in \mathcal{A}} \ U^*(s, a)\ \text{.}
\end{equation}

\vspace{0.3cm}

The novel methodology proposed by this research work to learn risk-sensitive decision-making policies based on the distributional RL approach is summarised as follows. Firstly, select any distributional RL algorithm that learns the full probability distribution of the random return $Z^{\pi}$. Secondly, leave the learning process unchanged except for action selection, which involves the maximisation of the utility function $U^{\pi}$ rather than the expected return $Q^{\pi}$. This adaptation is the single change to the distributional RL algorithm, occurring at two different locations within the algorithm: \textit{i.} the generation of new experiences by interacting with the environment, \textit{ii.} the learning of the random return $Z^{\pi}$ based on the distributional Bellman equation. However, this adaptation has no consequence on the random return $Z^{\pi}$ itself learnt by the distributional RL algorithm, only on the actions derived from that probability distribution. Algorithm \ref{AlgorithmRiskSensitive} details the proposed solution in a generic way, with the required modifications highlighted.

\section{Performance assessment methodology}
\label{SectionPerformanceAssessmentMethodology}

\subsection{Benchmark environments}
\label{SectionEnvironments}

\begin{figure}[H]
    \centering
    \begin{subfigure}[b]{0.28\textwidth}
        \centering
        \includegraphics[width=1\linewidth, trim={11cm 6.5cm 10.3cm 6cm}, clip]{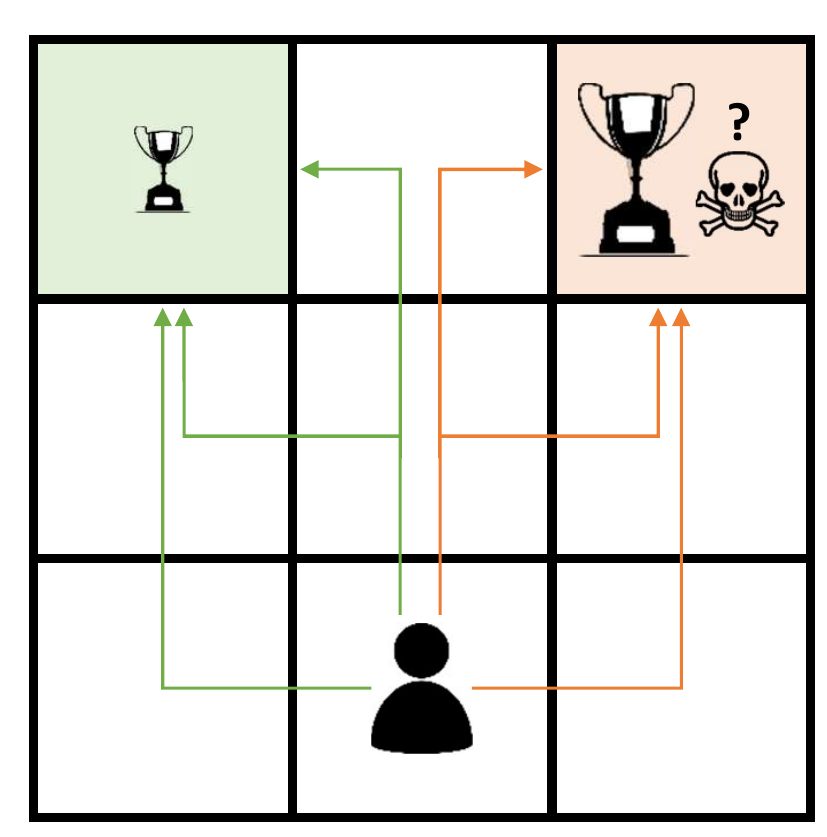}
        \caption{Risky rewards}
        \label{BenchmarkEnvironment1}
    \end{subfigure}
    \hfill
    \begin{subfigure}[b]{0.28\textwidth}
        \centering
        \includegraphics[width=1\linewidth, trim={11cm 6.5cm 10.3cm 4.3cm}, clip]{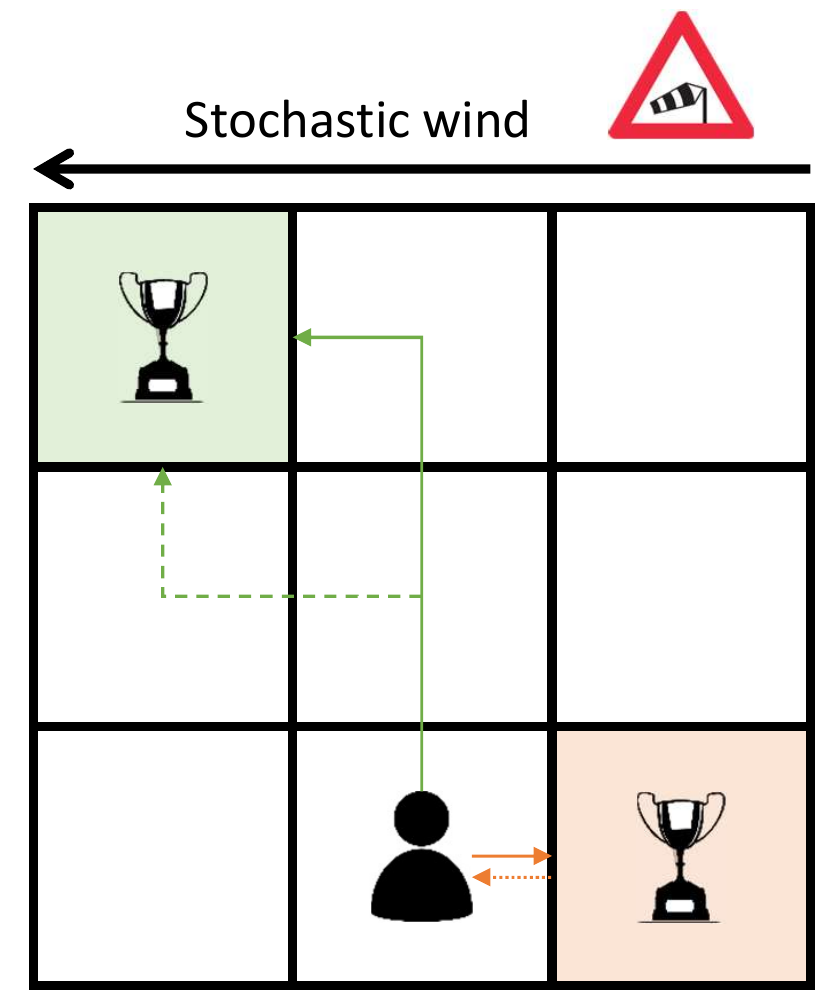}
        \caption{Risky transitions}
        \label{BenchmarkEnvironment2}
    \end{subfigure}
    \hfill
    \begin{subfigure}[b]{0.28\textwidth}
        \centering
        \includegraphics[width=1\linewidth, trim={11cm 6.5cm 10.3cm 4.3cm}, clip]{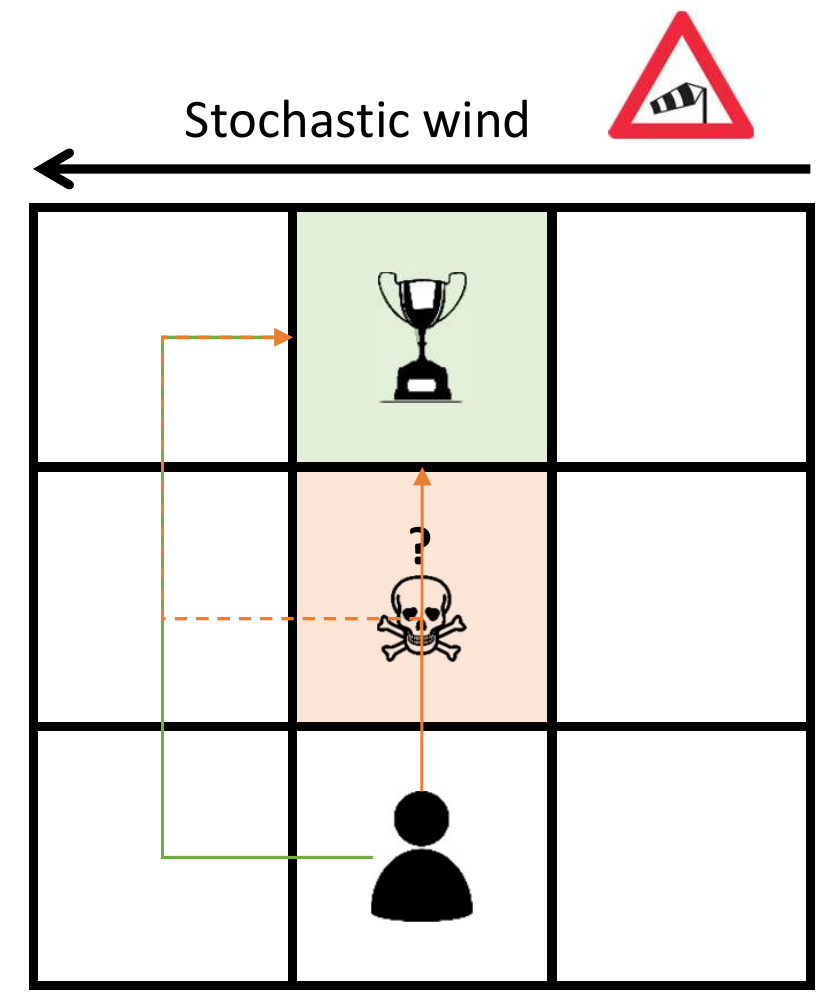}
        \caption{Risky grid world}
        \label{BenchmarkEnvironment3}
    \end{subfigure}
    \caption{Illustration of the benchmark environments introduced in this research work for the performance assessment of risk-sensitive decision-making policies. The optimal objective locations/paths in terms of risk mitigation and expected return maximisation are respectively highlighted in green and orange.}
    \label{FigureBenchmarkEnvironments}
\end{figure}

This research work introduces some novel benchmark environments in order to assess the soundness of the proposed methodology to design risk-sensitive policies based on the distributional RL approach. These environments consist of three toy problems that are specifically designed to highlight the importance of taking into consideration the risk for a decision-making policy. More precisely, the control problems are built in such a way that the optimal policy will differ depending on whether the objective is to solely maximise the expected performance or to also mitigate the risk. This is achieved by including relevant stochasticity in both the state transition function $p_T$ and reward function $p_R$. Moreover, the benchmark environments are designed relatively simple in order to ease the analysis and understanding of the decision-making policies learnt. This simplicity also ensures the accessibility of the experiments, since distributional RL algorithms generally require a considerable amount of computing power. Figure \ref{FigureBenchmarkEnvironments} illustrates these three benchmark environments, and highlights the optimal paths to be learnt depending on the objective pursued. For the sake of completeness, a thorough description of the underlying MDPs is provided in \ref{AppendixBenchmarkEnvironments}.\\

The first benchmark environment presented is named \textit{risky rewards}. It consists of a $3 \times 3$ grid world within which an agent has to reach one of two objective areas, that are equidistant from its fixed initial location. The difficulty of this control problem lies in the choice of the objective area to target, because of the stochasticity present in the reward function. Reaching the first objective area yields a reward with a lower value in expectation and a limited deviation from that average. On the contrary, reaching the second objective location yields a reward that is higher in expectation, at the cost of an increased risk.\\

The second benchmark environment studied is named \textit{risky transitions}. It consists of a $3 \times 3$ grid world within which an agent has to reach one of two objective areas as quickly as possible, in the presence of a stochastic wind. The agent is initially located in a fixed area that is very close to an objective, but the required move to reach it is in opposition to the wind direction. Following that path results in a reward that is higher in expectation, but there is a risk to be repeatedly countered by the stochastic wind. On the contrary, the longer path is safer but yields a lower reward on average.\\

The last benchmark environment presented is named \textit{risky grid world}. This control problem can be viewed as a combination of the two environments previously described since it integrates both stochastic rewards and transitions. It consists once again of a $3 \times 3$ grid world within which an agent, initially located in a fixed area, has to reach a fixed objective location as quickly as possible. To achieve that goal, three paths are available. The agent may choose the shortest path to the objective location that is characterised by a stochastic trap, or get around this risky situation by taking a significantly longer route. This bypass can be done from the left or from the right, another critical choice in terms of risk because of the stochastic wind. Once again, the optimal path is therefore dependent on the objective criterion to pursue.

\subsection{Risk-sensitive distributional RL algorithm analysed}
\label{SectionRLAlgorithm}

The distributional RL algorithm selected to assess the soundness of the methodology introduced for learning risk-sensitive decision-making policies is the \textit{Unconstrained Monotonic Deep Q-Network with Cramer} (UMDQN-C) \cite{Theate2021}. Basically, this particular distributional RL algorithm models the CDF of the random return in a continuous way by taking advantage of the Cramer distance for deriving the TD-error. Moreover, the probability distributions learnt are ensured to be valid thanks to the specific architecture exploited to model the random return: \textit{Unconstrained Monotonic Neural Network} (UMNN) \cite{Wehenkel2019}. The latter has been demonstrated to be a universal approximator of continuous monotonic functions, which is particularly convenient for representing CDFs. In practice, the UMDQN-C algorithm has been shown to achieve great results, both in terms of policy performance and in terms of random return probability distribution quality. This second feature clearly motivates the selection of this specific distributional RL algorithm to conduct the following experiments, since accurate random return probability distributions are required to properly estimate the risk. The reader can refer to the original research paper \cite{Theate2021} for more information about the UMDQN-C distributional RL algorithm.\\

As previously explained in Section \ref{SectionRisk}, the approach presented requires the choice of a function $\mathcal{R}_{\rho}$ for extracting risk features from the random return probability distribution $Z^{\pi}$. In the following experiments, the \textit{Value at Risk} (VaR) is adopted to estimate the risk. This choice is motivated by both the popularity of that risk measure in practice and by the efficiency of computation. Indeed, this quantity can be conveniently derived from the CDF of the random return learnt by the UMDQN-C algorithm.\\

In the next section presenting the results achieved, the risk-sensitive version of the UMDQN-C algorithm is denoted \textit{RS-UMDQN-C}. The detailed pseudo-code of that new risk-sensitive distributional RL algorithm can be found in \ref{AppendixAlgorithm}.\\

To conclude this section, ensuring the reproducibility of the results in a transparent way is particularly important to this research work. In order to achieve that, Table \ref{TableHyperparameters} provides a brief description of the key hyperparameters used in the experiments. Moreover, the entire experimental code is made publicly available at the following link: \url{https://github.com/ThibautTheate/Risk-Sensitive-Policy-with-Distributional-Reinforcement-Learning}.\\

\begin{table}[H]
  \caption{Description of the main hyperparameters used in the experiments.}
  \label{TableHyperparameters}
  \centering
  \begin{tabular}{lll}
    \toprule
    \textbf{Hyperparameter} & \textbf{Symbol} & \textbf{Value} \\
    \midrule
    DNN structure & - & $[128, 128]$ \\
    Learning rate & $L_r$ & $10^{-4}$ \\
    Deep learning optimiser epsilon & - & $10^{-5}$ \\
    Replay memory capacity & $C$ & $10^4$ \\
    Batch size & $N_e$ & $32$ \\
    Target update frequency & $N^-$ & $10^3$ \\
    Random return resolution & $N_z$ & $200$ \\
    Random return lower bound & $z_{\text{min}}$ & $-2$ \\
    Random return upper bound & $z_{\text{max}}$ & $+2$ \\
    Exploration $\epsilon$-greedy initial value & - & $1.0$ \\
    Exploration $\epsilon$-greedy final value & - & $0.01$ \\
    Exploration $\epsilon$-greedy decay & - & $10^4$ \\
    Risk coefficient & $\rho$ & $10\%$ \\
    Risk trade-off & $\alpha$ & $0.5$ \\
    \bottomrule
  \end{tabular}
\end{table}

\section{Results}
\label{SectionResults}

\subsection{Decision-making policy performance}
\label{SectionResultsPolicyPerformance}

To begin with, the performance achieved by the decision-making policies $\pi$ learnt has to be evaluated, both in terms of expected outcome and risk. For comparison purposes, the results obtained by the well-established DQN algorithm, a reference without any form of risk sensitivity, are presented alongside those of the newly introduced RS-UMDQN-C algorithm. It shall be mentioned that these two RL algorithms achieve very similar results when risk sensitivity is disabled ($\alpha = 1$ for the RS-UMDQN-C algorithm), as expected.\\

In the following, two analyses are presented. Firstly, the probability distribution of the cumulative reward of a policy $\pi$, denoted $S^{\pi} \in \mathcal{Z}$, is investigated. More precisely, the expectation $\mathbb{E}[\cdot]$, the risk function $\mathcal{R}_{\rho}[\cdot]$ and the utility function $U[\cdot]$ of that random variable $S^{\pi}$ are derived for each algorithm and compared. Secondly, this research work introduces a novel easy to interpret performance indicator $R_s \in [-1, 1]$ for evaluating the risk-sensitivity of the decision-making policies learnt, by taking advantage of the simplicity of the benchmark environments presented in Section \ref{SectionEnvironments}. In fact, it is made possible by the easy assessment from a human perspective of the relative riskiness of a path in the grid world environments studied. If the optimal path in terms of risk is chosen (green arrows in Figure \ref{FigureBenchmarkEnvironments}), a score $R_s = +1$ is awarded. On the contrary, the riskier path but optimal in expectation (orange arrows in Figure \ref{FigureBenchmarkEnvironments}) yields a score $R_s = -1$. If no objective nor trap areas are reached within the time allowed, a score $R_s = 0$ is delivered. Consequently, the evolution of this performance indicator provides valuable information about the convergence of the RL algorithms towards the different possible paths as well as about the stability of the learning process. Formally, let $\tau = \left\{s_t, a_t\right\}_{t\in[0, T]}$ with $s_t \in \mathcal{S}$ and $a_t \in \mathcal{A}$ be a trajectory defined over a time horizon $T < 10$ (ending with a terminal state, and subject to an upper bound), and let $\tau_+$ and $\tau_-$ respectively be the sets of trajectories associated to the green and orange paths in Figure \ref{FigureBenchmarkEnvironments}. Based on these definitions, the risk-sensitivity $R_s$ of a policy $\pi$ is a random variable that can be assessed via Monte Carlo as the following:
\begin{equation}
  R_s(\pi) =
    \begin{cases}
      +1 & \text{if $\pi$ produces trajectories $\left\{s_t, a_t\right\}_{t\in[0, T]} \in \tau_+$ with $a_t = \pi(s_t)$,}\\
      -1 & \text{if $\pi$ produces trajectories $\left\{s_t, a_t\right\}_{t\in[0, T]} \in \tau_-$ with $a_t = \pi(s_t)$,}\\
      0 & \text{otherwise.}
    \end{cases}
\end{equation}

\vspace{0.3cm}

\begin{figure}
    \centering
    \begin{subfigure}[b]{0.43\textwidth}
        \centering
        \includegraphics[width=1\linewidth, trim={0.7cm 0.0cm 2.5cm 1.8cm}, clip]{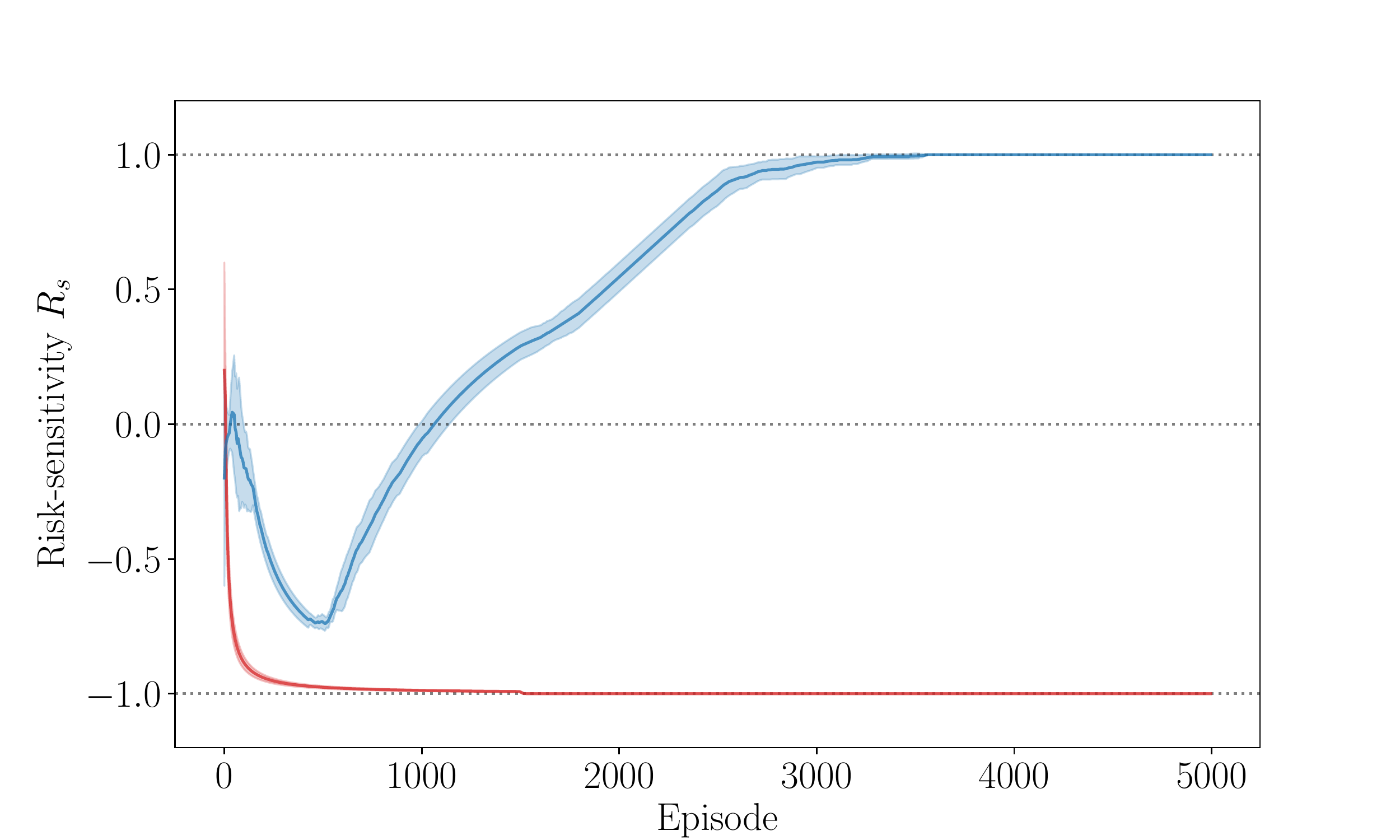}
        \caption{Risky rewards}
        \label{PolicyPerformanceBenchmarkEnvironment1}
    \end{subfigure}
    \hspace{1cm}
    \vspace{0.3cm}
    \begin{subfigure}[b]{0.43\textwidth}
        \centering
        \includegraphics[width=1\linewidth, trim={0.7cm 0.0cm 2.5cm 1.8cm}, clip]{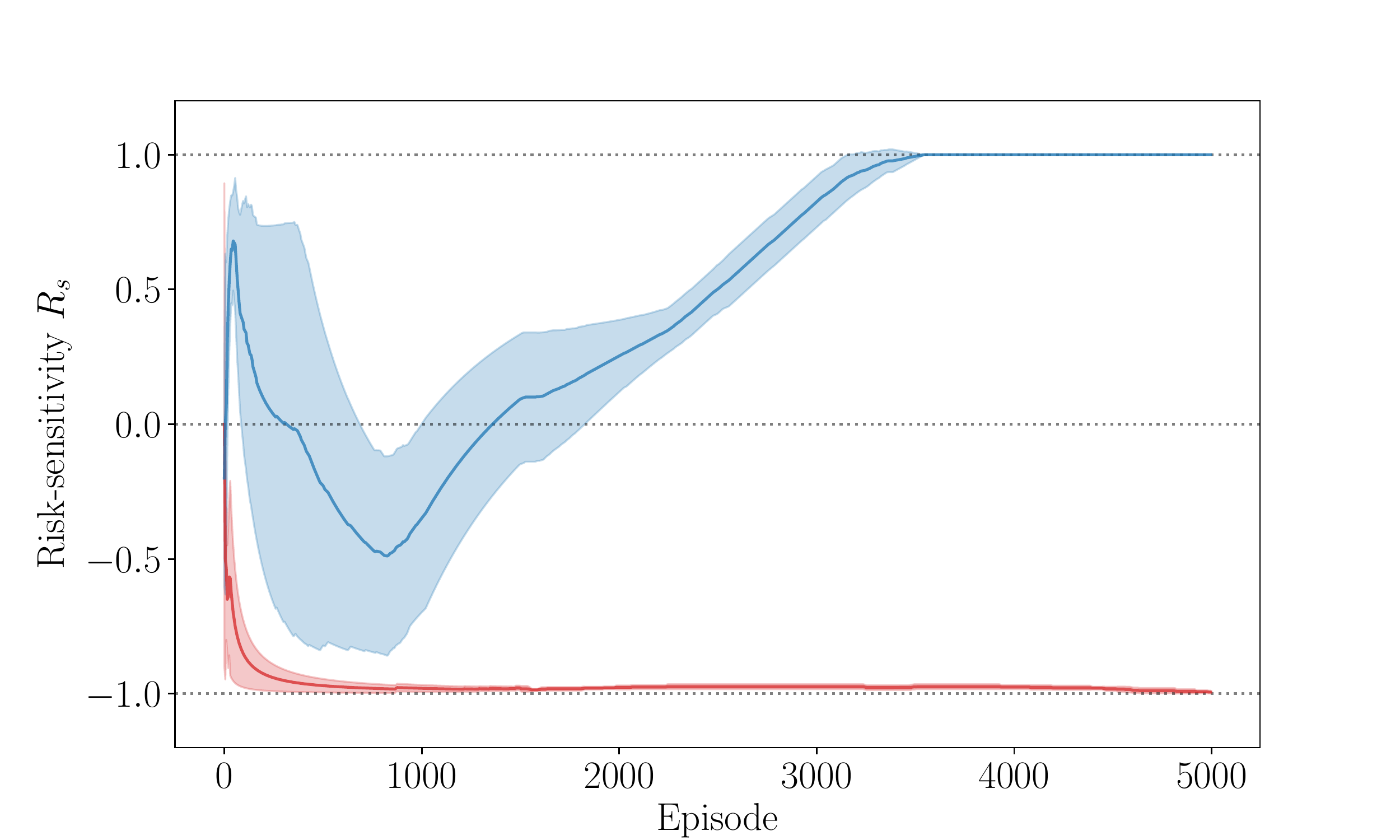}
        \caption{Risky transitions}
        \label{PolicyPerformanceBenchmarkEnvironment2}
    \end{subfigure}
    \begin{subfigure}[b]{0.43\textwidth}
        \centering
        \includegraphics[width=1\linewidth, trim={0.7cm 0.0cm 2.5cm 1.8cm}, clip]{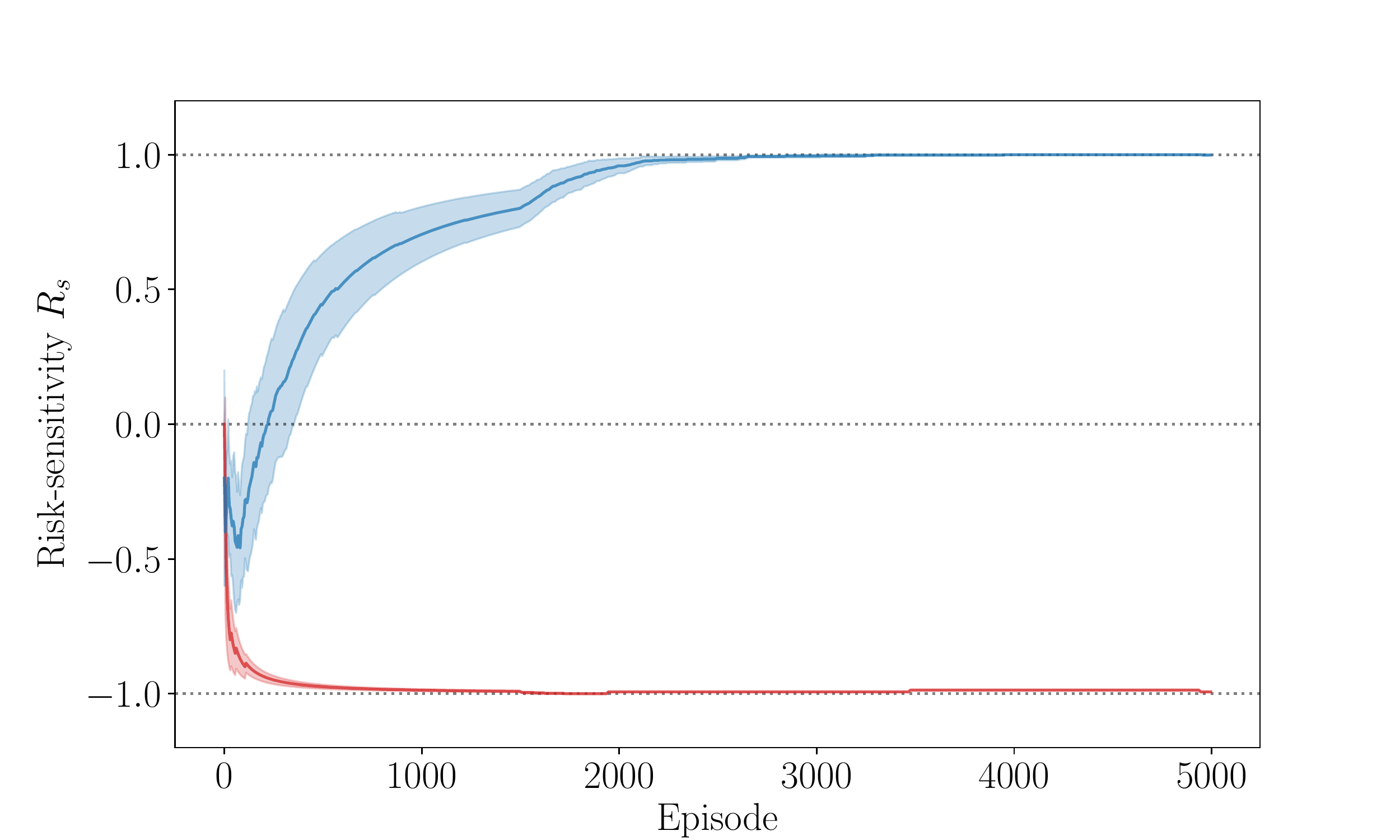}
        \caption{Risky grid world}
        \label{PolicyPerformanceBenchmarkEnvironment3}
    \end{subfigure}
    \hspace{3cm}
    \begin{subfigure}[b]{0.2\textwidth}
        \centering
        \includegraphics[width=1\linewidth, trim={15.5cm 7cm 2.7cm 2.0cm}, clip]{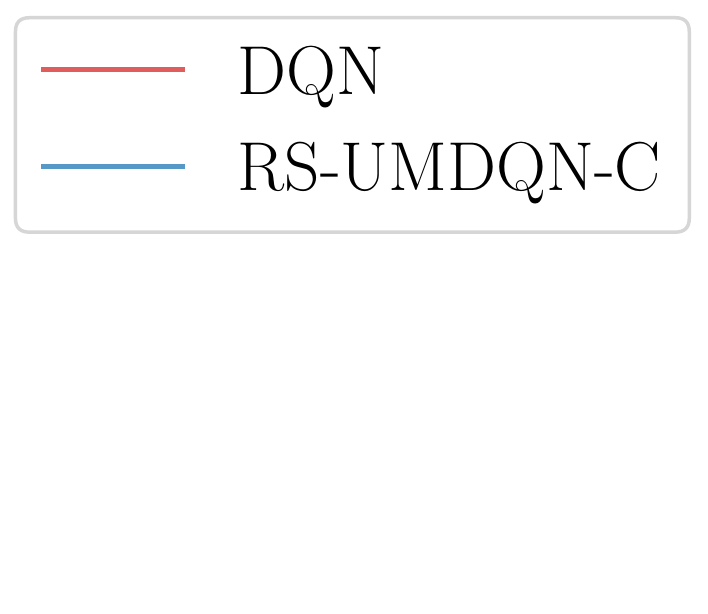}
        \label{LegendFigurePerformance}
    \end{subfigure}
    \hspace{1.7cm}
    \caption{Evolution of the risk-sensitivity performance indicator $R_s$ (expected value of the random variable) achieved by the decision-making policies $\pi$ learnt by both the DQN and RS-UMDQN-C algorithms during the training phase.}
    \label{FigurePolicyPerformance}
\end{figure}

The first results on policy performance are summarised in Table \ref{TablePolicyPerformance}, which compares the decision-making policies learnt by the DQN and RS-UMDQN-C algorithms both in terms of expected outcome and risk. The second results on policy performance are compiled in Figure \ref{FigurePolicyPerformance} plotting the evolution of the risk-sensitivity performance indicator $R_s$ during the training phase. It can be clearly observed from these two analyses that the proposed approach is effective in learning decision-making policies that are sensitive to the risk for relatively simple environments. As expected, the DQN algorithm yields policies that are optimal in expectation whatever the level of risk incurred. In contrast, the RS-UMDQN-C algorithm is able to leverage both expected outcome and risk in order to learn decision-making policies that produce a slightly lower expected return with a significantly lower risk level. This allows the proposed methodology to significantly outperform the risk-neutral RL algorithm of reference with respect to the performance indicator of interest $U[S^{\pi}]$ in Table \ref{TablePolicyPerformance}. Finally, it is also encouraging to observe from Figure \ref{FigurePolicyPerformance} that the learning process seems to be quite stable for simple environments, despite having to maximise a much more complicated function.

\begin{table}[H]
  \caption{Comparison of the expectation $\mathbb{E}[\cdot]$, the risk function $\mathcal{R}_{\rho}[\cdot]$ and the utility function $U[\cdot]$ of the cumulative reward $S^{\pi}$ achieved by the decision-making policies $\pi$ learnt by both the DQN and RS-UMDQN-C algorithms.}
  \label{TablePolicyPerformance}
  \centering
  \begin{tabular}{lcccccc}
    \toprule
    \multicolumn{1}{c}{\multirow{2}{*}{\textbf{Benchmark environment}}} & \multicolumn{3}{c}{\textbf{DQN}} & \multicolumn{3}{c}{\textbf{RS-UMDQN-C}} \\
    \cmidrule(r){2-7}
    & $\mathbb{E}\left[ S^{\pi} \right]$ & $\mathcal{R}_\rho\left[ S^{\pi} \right]$ & $U\left[ S^{\pi} \right]$ & $\mathbb{E}\left[ S^{\pi} \right]$ & $\mathcal{R}_\rho\left[ S^{\pi} \right]$ & $U\left[ S^{\pi} \right]$ \\
    \midrule
    Risky rewards & \textbf{0.3} & -1.246 & -0.474 & 0.1 & \textbf{-0.126} & \textbf{-0.013} \\
    Risky transitions & \textbf{0.703} & 0.118 & 0.411 & 0.625 & \textbf{0.346} & \textbf{0.485} \\
    Risky grid world & \textbf{0.347} & -1.03 & -0.342 & 0.333 & \textbf{0.018} & \textbf{0.175} \\
    \bottomrule
  \end{tabular}
\end{table}

\subsection{Probability distribution visualisation}
\label{SectionResultsDistribution}

A core advantage of the proposed solution is the improved interpretability of the resulting decision-making process. Indeed, understanding and motivating the decisions outputted by the learnt policy $\pi$ is greatly facilitated by the access to the probability distributions of the random return jointly learnt. In addition, the analysis and comparison of the value, risk and utility functions ($Q^{\pi}$, $R^{\pi}$ and $U^{\pi}$) associated with different actions provide a valuable summary about the decision-making process, but also about the control problem itself. Such an analysis may be particularly important to correctly tune the risk trade-off parameter $\alpha$ according to the user's risk aversion.\\

\begin{figure}
    \centering
    \begin{subfigure}[b]{0.495\textwidth}
        \centering
        \includegraphics[width=1\linewidth, trim={1.0cm 1.1cm 1.9cm 2.0cm}, clip]{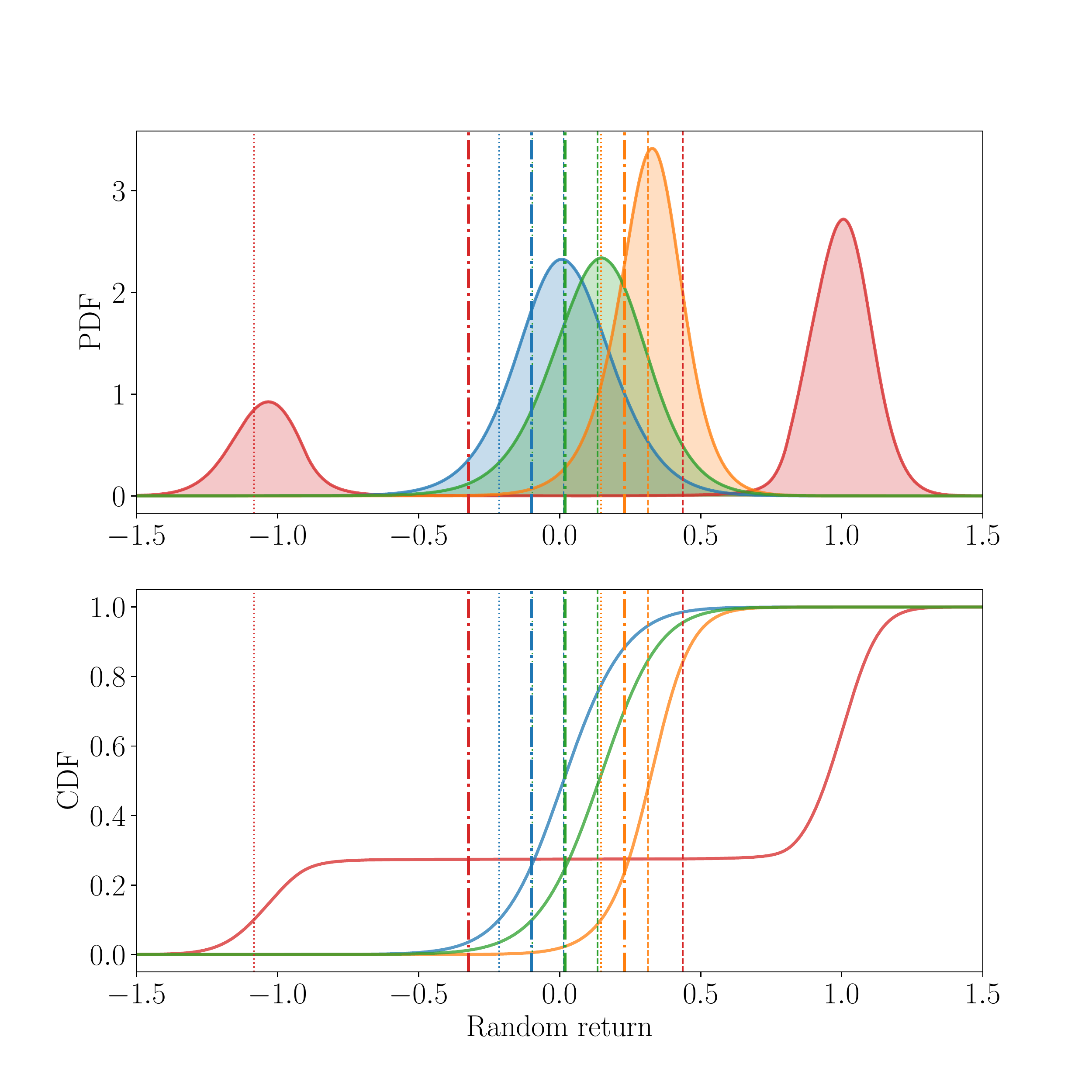}
        \caption{Risky rewards}
        \label{DistributionVisualisationBenchmarkEnvironment1}
    \end{subfigure}
    \hfill
    \begin{subfigure}[b]{0.495\textwidth}
        \centering
        \includegraphics[width=1\linewidth, trim={1.0cm 1.1cm 1.9cm 2.0cm}, clip]{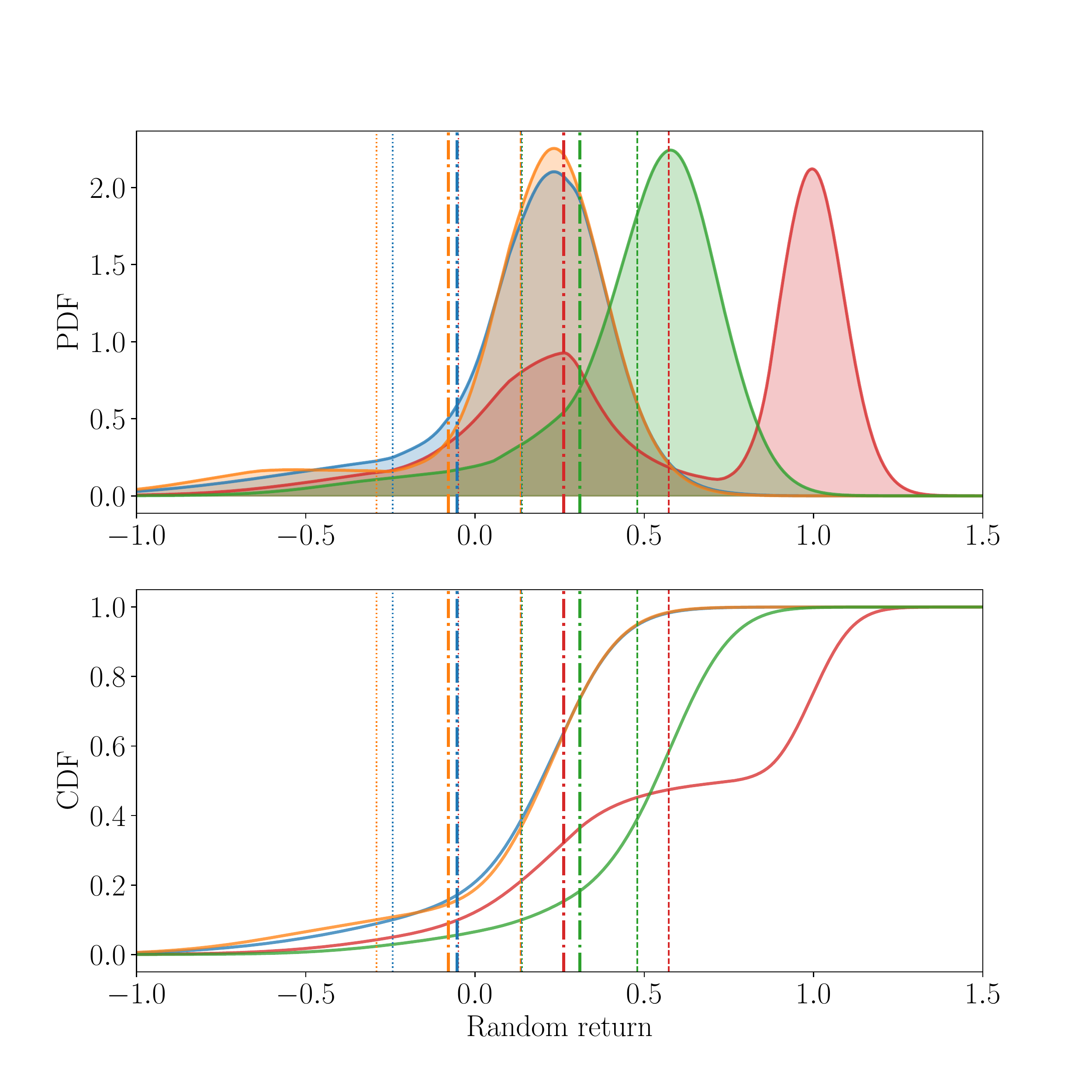}
        \caption{Risky transitions}
        \label{DistributionVisualisationBenchmarkEnvironment2}
    \end{subfigure}
    \hfill
    \begin{subfigure}[b]{0.495\textwidth}
        \centering
        \includegraphics[width=1\linewidth, trim={1.0cm 1.1cm 1.9cm 2.0cm}, clip]{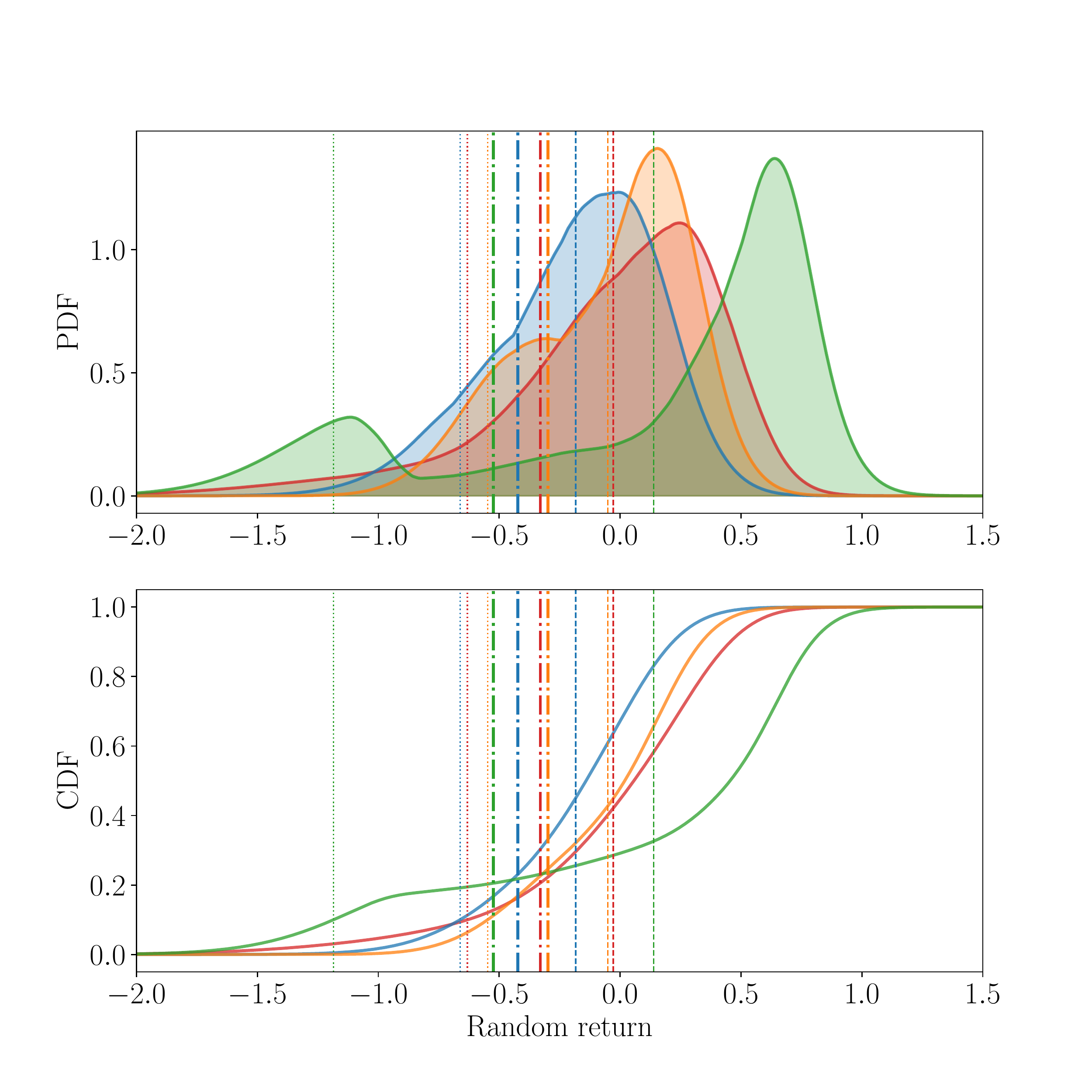}
        \caption{Risky grid world}
        \label{DistributionVisualisationBenchmarkEnvironment3}
    \end{subfigure}
    \hfill
    \begin{subfigure}[b]{0.495\textwidth}
        \centering
        \includegraphics[width=1\linewidth, trim={1.5cm 2.0cm 1.5cm 4.0cm}, clip]{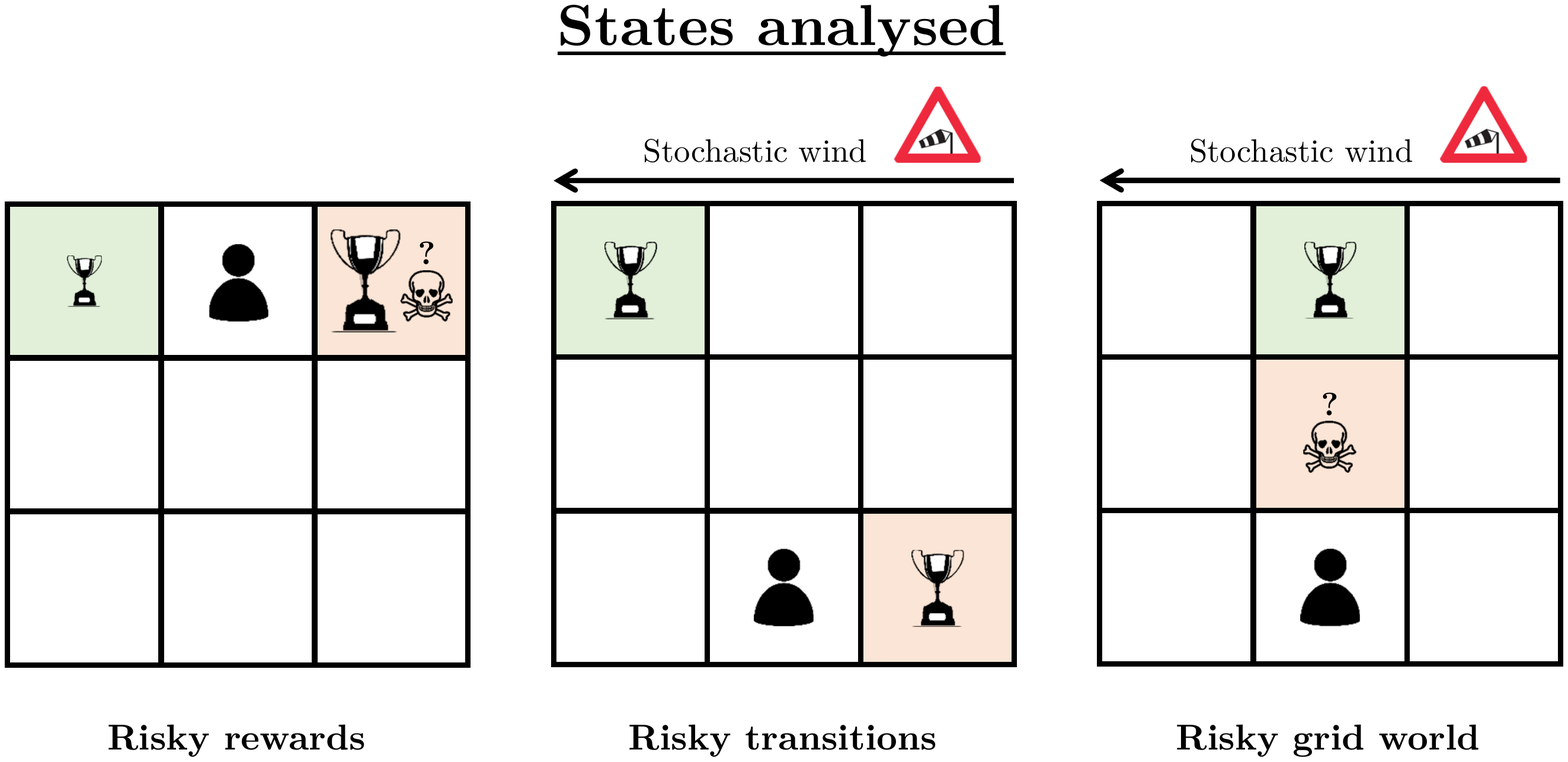}
        \includegraphics[width=0.75\linewidth, trim={9.5cm 15cm 2.8cm 3.3cm}, clip]{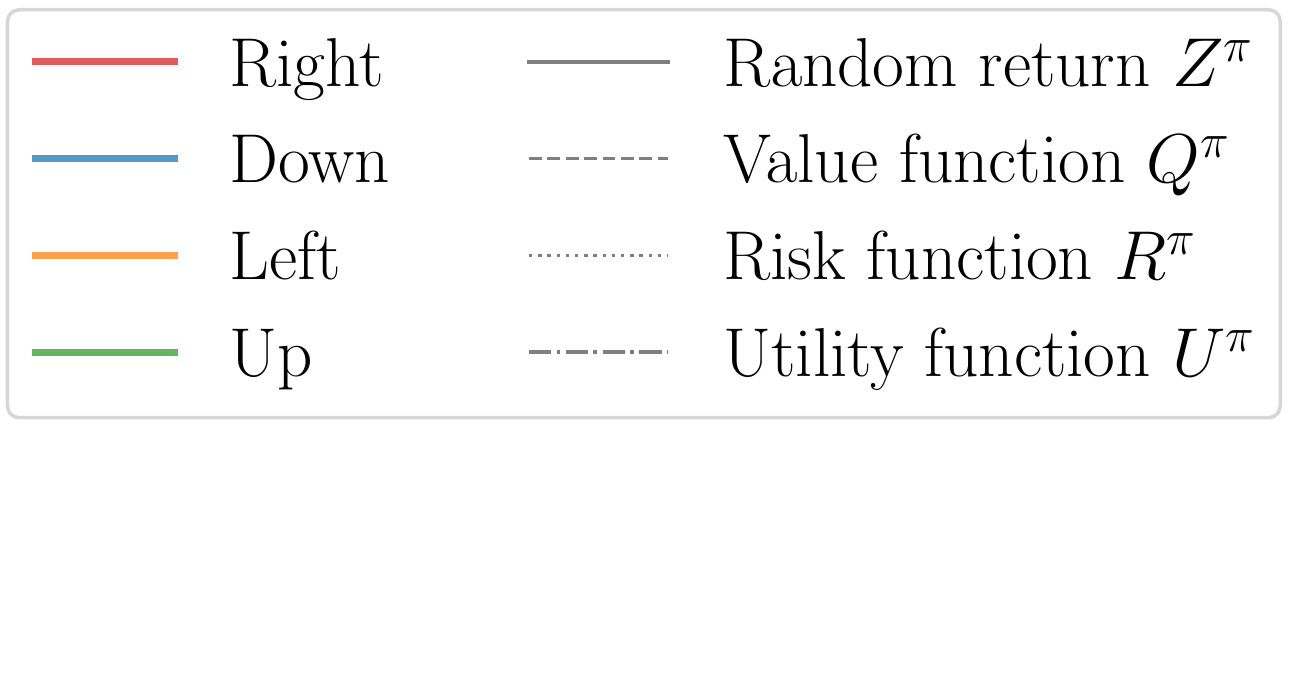}
        \label{LegendFigureDistribution}
    \end{subfigure}
    \caption{Visualisation of the random return probability distributions $Z^{\pi}$ learnt by the RS-UMDQN-C algorithm for typical states of the benchmark environments, together with the value, risk and utility functions derived ($Q^{\pi}$, $R^{\pi}$ and $U^{\pi}$).}
    \label{FigureDistributionVisualisation}
\end{figure}

As an illustration, Figure \ref{FigureDistributionVisualisation} demonstrates some random return probability distributions $Z^{\pi}$ that are learnt by the RS-UMDQN-C algorithm. More precisely, a single relevant state is selected for analysis for each benchmark environment. The selection is based on the importance of the next decision in following a clear path, either maximising the expected outcome or mitigating the risk. Firstly, it can be observed that the risk-sensitive distributional RL algorithm does manage to accurately learn the probability distributions of the random return, qualitatively from a human perspective. In particular, the multimodality purposely designed to create risky situations appears to be well preserved. Such a result is particularly encouraging since this feature is essential to the success of the proposed methodology. Indeed, it ensures the accurate estimation of the risk as defined in Section \ref{SectionRisk}. This observation is in line with the findings of the research paper \cite{Theate2021} introducing the UMDQN algorithm, and suggests that the solution introduced to achieve risk-sensitivity does not alter too much the properties of the original distributional RL algorithm. Secondly, as previously explained, Figure \ref{FigureDistributionVisualisation} highlights the relevance of each function introduced ($Q^{\pi}$, $R^{\pi}$ and $U^{\pi}$) for making and motivating a decision. Their analysis truly contributes to the understanding of the potential trade-off between expected performance maximisation and risk mitigation for a given decision-making problem, as well as the extent to which different values of the important parameter $\alpha$ leads to divergent policies.

\section{Conclusion}
\label{SectionConclusion}

The present research work introduces a straightforward yet efficient solution to learn risk-sensitive decision-making policies based on the distributional RL approach. The proposed methodology presents key advantages. Firstly, it is perfectly compatible with any distributional RL algorithm, and requires only minimal modification to the original algorithm. Secondly, the simplicity of the approach contributes to the interpretability and ease of analysis of the resulting risk-sensitive policies, a particularly important feature to avoid black-box machine learning models. Lastly, the solution presented allows to cover the complete potential trade-off between expected outcome maximisation and risk minimisation. The first experiments performed on three relevant toy problems yield promising results, which may be viewed as a proof of concept for the accessible and practical solution introduced.\\

Some interesting leads can be suggested as future work. Firstly, the research conducted is exclusively empirical and does not study any theoretical guarantees about the resulting risk-sensitive distributional RL algorithms. Among others, the study of the convergence of these algorithms would be a relevant future research direction. Secondly, building on the promising results achieved, the solution presented should definitely be evaluated on more complex environments, for which the risk should ideally be mitigated. Lastly, the approach could be extended to not only mitigate the risk but also to completely discard actions that would induce an excessive level of risk, in order to increase compliance with the objective criterion originally defined in Section \ref{SectionObjective}.

\section*{Acknowledgements}

Thibaut Théate is a Research Fellow of the F.R.S.-FNRS, of which he acknowledges the financial support.

\bibliography{manuscript}

\clearpage

\appendix

\section{Benchmark environments}
\label{AppendixBenchmarkEnvironments}

\paragraph{\textbf{Risky rewards environment}} The underlying MDP can be described as follows:
\begin{itemize}
    \item [$\bullet$] $\mathcal{S} \in \{0, 1, 2 \} \times \{0, 1, 2 \}$, a state $s$ being composed of the two coordinates of the agent within the grid,
    \item [$\bullet$] $\mathcal{A} = \{\texttt{RIGHT},\ \texttt{DOWN},\ \texttt{LEFT},\ \texttt{UP}\}$, with an action $a$ being a moving direction,
    \item [$\bullet$] $p_R(r|s, a) \sim \mathcal{N}(\mu,\sigma^{2})$ where:
        \begin{itemize}
            \item $\mu = 0.3$  and $\sigma = 0.1$ if the agent reaches the first objective location (terminal state),
            \item $\mu = 1.0$ and $\sigma = 0.1$ with a 75\% chance, and $\mu = -1.0$ and $\sigma = 0.1$ with a 25\% chance if the agent reaches the second objective location (terminal state),
            \item $\mu = -0.1$  and $\sigma = 0.1$ otherwise,
        \end{itemize}
    \item [$\bullet$] $p_T(s'|s, a)$ associates a 100\% chance to move once in the chosen direction, while keeping the agent within the $3 \times 3$ grid world (crossing a border is not allowed),
    \item [$\bullet$] $p_0$ associates a probability of 1 to the state $s = [1,0]$, which is the position of the agent in Figure \ref{FigureBenchmarkEnvironments},
    \item [$\bullet$] $\gamma = 0.9$.\\
\end{itemize}

\paragraph{\textbf{Risky transitions environment}} The underlying MDP can be described as the following:
\begin{itemize}
    \item [$\bullet$] $\mathcal{S} \in \{0, 1, 2 \} \times \{0, 1, 2 \}$, a state $s$ being composed of the two coordinates of the agent within the grid,
    \item [$\bullet$] $\mathcal{A} = \{\texttt{RIGHT},\ \texttt{DOWN},\ \texttt{LEFT},\ \texttt{UP}\}$, with an action $a$ being a moving direction,
    \item [$\bullet$] $p_R(r|s, a) \sim \mathcal{N}(\mu,\sigma^{2})$ where:
        \begin{itemize}
            \item $\mu = 1.0$ and $\sigma = 0.1$ if the agent reaches one of the objective locations (terminal state),
            \item $\mu = -0.3$  and $\sigma = 0.1$ otherwise,
        \end{itemize}
    \item [$\bullet$] $p_T(s'|s, a)$ associates a 100\% chance to move once in the chosen direction AND a 50\% chance to get pushed once to the left by the stochastic wind, while keeping the agent within the $3 \times 3$ grid world,
    \item [$\bullet$] $p_0$ associates a probability of 1 to the state $s = [1,0]$, which is the position of the agent in Figure \ref{FigureBenchmarkEnvironments},
    \item [$\bullet$] $\gamma = 0.9$.\\
\end{itemize}

\paragraph{\textbf{Risky grid world environment}} The underlying MDP is described as follows:
\begin{itemize}
    \item [$\bullet$] $\mathcal{S} \in \{0, 1, 2 \} \times \{0, 1, 2 \}$, a state $s$ being composed of the two coordinates of the agent within the grid,
    \item [$\bullet$] $\mathcal{A} = \{\texttt{RIGHT},\ \texttt{DOWN},\ \texttt{LEFT},\ \texttt{UP}\}$, with an action $a$ being a moving direction,
    \item [$\bullet$] $p_R(r|s, a) \sim \mathcal{N}(\mu,\sigma^{2})$ where:
        \begin{itemize}
            \item $\mu = 1.0$  and $\sigma = 0.1$ if the agent reaches the objective location (terminal state),
            \item $\mu = -0.2$ and $\sigma = 0.1$ with a 75\% chance, and $\mu = -2.0$ and $\sigma = 0.1$ with a 25\% chance if the agent reaches the stochastic trap location (terminal state),
            \item $\mu = -0.2$  and $\sigma = 0.1$ otherwise,
        \end{itemize}
    \item [$\bullet$] $p_T(s'|s, a)$ associates a 100\% chance to move once in the chosen direction AND a 25\% chance to get pushed once to the left by the stochastic wind, while keeping the agent within the $3 \times 3$ grid world,
    \item [$\bullet$] $p_0$ associates a probability of 1 to the state $s = [1,0]$, which is the position of the agent in Figure \ref{FigureBenchmarkEnvironments},
    \item [$\bullet$] $\gamma = 0.9$.
\end{itemize}

\newpage

\section{RS-UMDQN-C algorithm}
\label{AppendixAlgorithm}

Algorithm \ref{RS-UMDQN-C} presents the novel \textit{Risk-Sensitive Unconstrained Monotonic Deep Q-Network with Cramer} algorithm (RS-UMDQN-C). Basically, this new RL algorithm is the result of the application of the methodology described in Algorithm \ref{AlgorithmRiskSensitive} to the UMDQN-C algorithm thoroughly introduced in \cite{Theate2021}.

\begin{algorithm*}
\caption{RS-UMDQN-C algorithm}
\small
\begin{algorithmic} 
\STATE Initialise the experience replay memory $M$ of capacity $C$.
\STATE Initialise the main UMNN weights $\theta$ (Xavier initialisation).
\STATE Initialise the target UMNN weights $\theta^- = \theta$.
\FOR{episode = 0 \TO $N$}
    \FOR{$t = 0$ \TO $T$, or until episode termination}
        \STATE Acquire the state $s$ from the environment $\mathcal{E}$.
        \STATE With probability $\epsilon$, select a random action $a \in \mathcal{A}$.
        \STATE Otherwise, select $a = \argmax_{a' \in \mathcal{A}} U(s, a'; \theta)$, with $U(s, a'; \theta) = \alpha \ \mathbb{E}\left[ Z(s, a'; \theta) \right] \ + \ (1 - \alpha) \ \text{VaR}_{\rho} \left[ Z(s, a'; \theta) \right]$.
        \STATE Interact with the environment $\mathcal{E}$ with action $a$ to get the next state $s'$ and the reward $r$.
        \STATE Store the experience $e = (s, a, r, s')$ in the experience replay memory $M$.
        \STATE Randomly sample from $M$ a minibatch of $N_e$ experiences $e_i = (s_i, a_i, r_i, s_{i}^{'})$.
        \STATE Derive a discretisation of the domain $\mathcal{X}$ by sampling $N_z$ returns $z \sim \mathcal{U}([z_{\text{min}}, z_{\text{max}}])$.
        \FOR{$i = 0$ \TO $N_e$}
            \FORALL{$z \in \mathcal{X}$}
                \IF{$s_{i}^{'}$ is terminal}
                    \STATE Set $y_i(z) = \begin{cases}
                                                0 & \text{if } z < r_i, \\
                                                1 & \text{otherwise.}
                                            \end{cases}$
                \ELSE
                    \STATE Set $y_i(z) = Z\left(\frac{z - r_i}{\gamma} \bigg| s_{i}^{'}, \argmax_{a_{i}^{'} \in \mathcal{A}} U(s_{i}^{'}, a_{i}^{'}; \theta^-); \theta^-\right)$.
                \ENDIF
            \ENDFOR
        \ENDFOR
        \STATE Compute the loss $\mathcal{L}_C(\theta) = \sum_{i=0}^{N_e} \left(\sum_{z \in \mathcal{X}} \left(y_i(z) - Z(z|s_i, a_i; \theta)\right)^2\right)^{1/2}$.
        \STATE Clip the resulting gradient in the range $[0, 1]$.
        \STATE Update the main UMNN parameters $\theta$ using the ADAM optimiser with learning rate $L_r$.
        \STATE Update the target UMNN parameters $\theta^- = \theta$ every $N^-$ steps.\\
        \STATE Anneal the $\epsilon$-greedy exploration parameter $\epsilon$.
    \ENDFOR
\ENDFOR
\end{algorithmic} 
\label{RS-UMDQN-C}
\end{algorithm*}

\end{document}